\documentclass[a4paper]{article}
\usepackage{cite}
\usepackage{url}
\usepackage{enumitem}
\usepackage{hyperref}

\newcommand{\footref}[1]{%
    $^{\ref{#1}}$%
}

\usepackage{INTERSPEECH2021}

\title{Combining Spectral and Self-Supervised Features for Low Resource Speech Recognition and Translation}
\name{Dan Berrebbi$^1$, Jiatong Shi$^1$, Brian Yan$^1$, Osbel López-Francisco$^2$, Jonathan D. Amith$^3$, Shinji~Watanabe$^1$}
\address{
  $^1$Language Technologies Institute, Carnegie Mellon University\\
  $^2$Universidad Nacional Autónoma de México, Iztacala \\
  $^3$Dept. of Anthropology, Gettysburg College}
\email{dberrebb@andrew.cmu.edu, jiatongs@andrew.cmu.edu}

\begin{document}

\maketitle

\begin{abstract}
Self-Supervised Learning (SSL) models have been successfully applied in various deep learning-based speech tasks, particularly those with a limited amount of data. However, the quality of SSL representations depends highly on the relatedness between the SSL training domain(s) and the target data domain. 
On the contrary, spectral feature (SF) extractors such as log Mel-filterbanks are hand-crafted non-learnable components, and could be more robust to domain shifts. The present work examines the assumption that combining non-learnable SF extractors to SSL models is an effective approach to low resource speech tasks.
We propose a learnable and interpretable framework to combine SF and SSL representations. The proposed framework outperforms significantly both baseline and SSL models on Automatic Speech Recognition (ASR) and Speech Translation (ST) tasks on three low resource datasets. We additionally design a mixture of experts based combination model. This last model reveals that the relative contribution of SSL models over conventional SF extractors is very small in case of domain mismatch between SSL training set and the target language data.


\end{abstract}

\section{Introduction}

End-to-end models based on deep learning have demonstrated their superiority over conventional hidden Markov-based models on speech tasks for some corpora \cite{chiu2018state, karita2019comparative, pham2019very, guo2021recent}. End-to-end models could be beneficial to low resource speech tasks because these models: (1) alleviate the need of language specific resources such as lexicons \cite{grenoble2011handbook, zahrer2020towards, shi-etal-2021-leveraging}. (2) can be trained multilingually to facilitate cross-lingual transfers between high resource and low resource languages through shared architecture and weights \cite{8639655}. On the other hand, end-to-end models can perform poorly when the training data is limited \cite{rwth-libri-2019} and low resource scenarios often introduce a language-mismatch with the data used to train powerful self-supervised learning (SSL) representations \cite{tsai2022superbsg}.


One direction towards mitigating these low-resource issues is to incorporate knowledge from several languages into multilingual end-to-end models \cite{watanabe2017language, toshniwal2018multilingual, Kannan2019}.
When there is no training data available for the target languages, these systems can be even applied in a zero-shot manner \cite{li2020universal, yan2021differentiable, xu2021simple}.
Fortunately, many languages have small amounts of data which can be used to fine-tune large-scale multilingual models towards target languages, resulting in further improvements \cite{hou20_interspeech, pratap20c_interspeech, adams2019massively, li2021scaling}.

Another direction is to use self-supervised learning models trained on large untranscribed corpora as front-end feature extractors, replacing conventional spectral features (SF) such as log Mel-filterbanks coefficients (FBANK) \cite{yi2020applying, wu20g_interspeech, baevski2020wav2vec, n21_interspeech, chang2021exploration, liu2021tera}.  During their unsupervised training, SSL models \cite{DBLP:HuBERT, DBLP:wav2vec,DBLP:wav2vec2,chen2021wavlm} learn their own feature extraction modules and are totally free of SF at fine-tuning time. As these models achieve state of the art on numerous speech 
tasks and significantly outperform models with more supervision, the effectiveness of SF on low resource tasks is increasingly questioned. 

The majority of SSL models are trained exclusively using English speech. Although these approaches have shown improvements, even when domain mismatches occur (such as language or audio conditions \cite{sanabria2022measuring}), performance depends on the relatedness between the SSL training domain and the target language one \cite{conneau2019unsupervised}. SSL first layers output representations tend to be quite similar to SF according to a canonical correlation analysis \cite{andrew2013deep} of Wav2vec2 \cite{DBLP:wav2vec2} from Pasad et al. \cite{DBLP:layerwiseToyota}. In contrast, the last layers are likely to be more corpus or domain-specific, which should be randomly initialized at fine-tuning time \cite{DBLP:layerwiseToyota}. Therefore, we assume that  SSL representations are potentially more hurted by domain shifts than SF-based systems are. SF are domain and language agnostic and their use in multilingual models has demonstrated that they enable strong cross-lingual transfers \cite{8639655}. It is then legitimate to assume that a model leveraging both SF and SSL representations would lead to strong performances on low resource speech scenarios.

In the present work, we examine this assumption by building a framework that enables combining SF and SSL representations through learnable fusions. We propose linear, convolutional and co-attention based combinations. Those methods obtain a relative diminution of $19.3\%$ Character Error Rate (CER), averaged on two ASR datasets, and a gain of 1.0 BLEU, on an ST dataset, over the SSL baseline model, while having less than 0.01\% additional parameters. We further propose a mixture of experts \cite{MoE} based technique in order to better interpret the roles and complementarities of SF and SSL components.\footnote{Our code is released on \href{https://github.com/espnet/espnet}{ESPnet} \cite{watanabe2018espnet}} Finally the proposed framework is evaluated on Totonac, a Mexican endangered language, and we release the first publicly available annotated speech corpus of this language.\footnote{\url{http://www.openslr.org/107/}}

\section{Speech Representations \label{FBANK_sec}}


\par \textbf{Spectral Features:} Machine learning based speech analytics require the extraction of feature vectors from raw analog waveforms. Log Mel-filterbanks features (FBANK), conventionally used for supervised speech processing tasks, are perceptually inspired by human hearing. These features sample and quantize the analog waveform, apply pre-emphasis to boost high frequency energies, undergo a discrete Fourier transform (DFT), and finally passed through Mel filter banks. It is worth noting that the DFT operation is linear and could be learned during model training but the system may fail to learn it due to its high complexity, especially if only small amounts of data are available.





\noindent \textbf{Self-Supervised Learning features :}  While FBANK are hand-crafted features inspired by the human perception of speech, SSL features learn latent representations derived from large amounts of unlabeled data. After training the SSL model, often referred to as pre-training, a fine-tuning phase is conducted with a task-specific labeled data set. The key idea is that unlabeled data contains valuable information and is far more abundant than labeled data in any domain. This paradigm leads to general-purpose speech representation, suitable for speech processing tasks \cite{tsai2022superbsg}. 


\section{Proposed Approaches}

\subsection{Feature extraction \label{sec_feature_extraction}}


Let $S$ be a sampled and quantized raw waveform of one utterance.
We note $f_{\text{SF}}(S)$ and $f_{\text{SSL}}(S)$ the features extracted from $S$ by spectral feature extractors and SSL models (respectively SF and SSL in formulas). We note $T_{\text{SF}}$ and $T_{\text{SSL}}$ the number of frames of the utterance, while $D_{\text{SF}}$ and $D_{\text{SSL}}$ denote dimensions of the features extracted by $f_{\text{SF}}$ and $f_{\text{SSL}}$. We obtain,


\vspace{-12pt}
\begin{equation}
    f_{\text{i}}(S) = (f^{t}_{\text{i}}(S) \in \mathbb{R} ^{D_{\text{i}}}| t=1, \cdots, T_{\text{i}}), \: i\in \{\text{SF, SSL}\}
\end{equation} 
Additional linear projection and reshaping is applied over SF and SSL features to allow a same feature dimension $D=D_{\text{SF}}$ and number of frames $T=T_{\text{SSL}}$. 
For the dimension, we choose to project SSL features into SF space and not the inverse in order to decrease the number of parameters (as $D_{\text{SF}} < D_{\text{SSL}}$) for efficiency purposes. For the number of frames, as we use a frame-shift two times longer for SSL than for SF, we downsample (through linear projection and reshaping) the SF features to get a common number of frames $T=T_{\text{SSL}}$. We now have $f_{\text{SF}}(S)\in  \mathbb{R}^{T\times D}$ and $f_{\text{SSL}}(S)\in  \mathbb{R}^{T\times D}$. Our goal is to combine $f_{\text{SF}}(S)$ and $f_{\text{SSL}}(S)$ in order to get the best model for low resource tasks. 

\vspace{-4pt}
\subsection{Learnable combinations \label{sec_feature_comb}}
\vspace{-2pt}
We first propose a general framework of using learnable transformations (concatenation, convolutional, and co-attention \cite{coattention} mechanisms) for combining those features. Such learnable fusions have previously been employed in various multisource/multimodal applications\cite{libovicky-helcl-2017-attention, hori2017attention}.
The framework is formulated as follows, where $f_{\text{FUSE}}(S)$ is the resultant features:
\vspace{-5pt}
\begin{equation}
\vspace{-2pt}
    f_{\text{FUSE}}(S) = \textsc{Linear(Transform}(f_{\text{SF}}(S),f_{\text{SSL}}(S))
    \label{eq_transform}
\end{equation}

\begin{figure}[!ht]
    \centering
    \includegraphics[scale=.31]{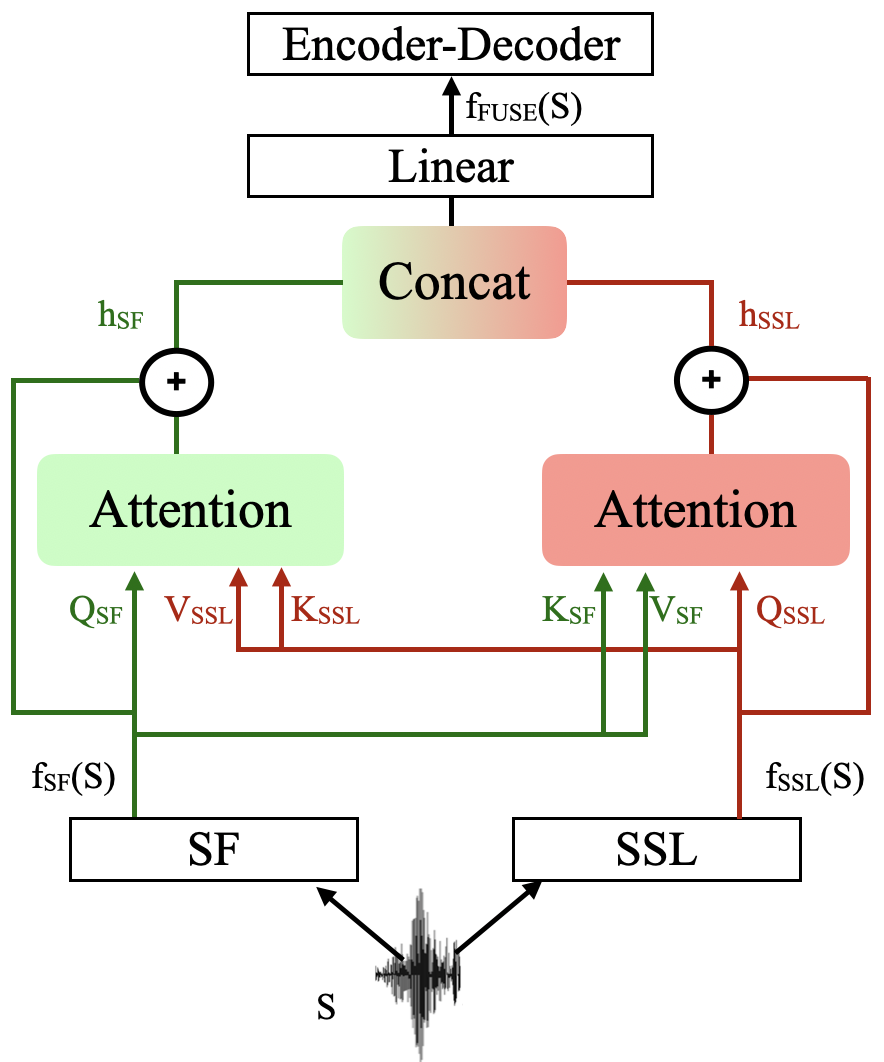}
    \vspace{-5pt}
    \caption{Architecture of our proposed co-attention based fusion. Raw signal S is passed through SF and SSL feature extractors. The extracted features, $f_{\text{SF}}(S)$ and $f_{\text{SSL}}(S)$, attend to each other through two distinct attention mechanisms. Output features are then concatenated, projected and passed to the speech model.}
    \label{fig:architecture_coatt}
    \vspace{-15pt}
\end{figure}

\noindent With \textsc{Transform} being a concatenation, a convolution or a \textbf{co-attention based fusion}.  We will dive into more details about Eq.~\eqref{eq_transform} for the proposed co-attention fusion method, which is illustrated in Fig.~\ref{fig:architecture_coatt}.

\noindent Let $W_{\text{SF}}^{\text{Q}}$, $W_{\text{SF}}^{\text{K}}$, $W_{\text{SF}}^{\text{V}}$, $W_{\text{SSL}}^{\text{Q}}$, $W_{\text{SSL}}^{\text{K}}$ and $W_{\text{SSL}}^{\text{V}}$ be six learnable matrices of $\mathbb{R}^{D\times D}$. 
We use classical attention notation \cite{vaswani} in Eq.~\eqref{eq_query}. For i$\in$\{SF, SSL\}, we note,
\begin{equation}
\vspace{-3pt}
Q_{\text{i}} = f_{\text{i}}(S)W_{\text{i}}^{\text{Q}}, \quad K_{\text{i}} = f_{\text{i}}(S)W_{\text{i}}^{\text{K}}, \quad V_{\text{i}} = f_{\text{i}}(S)W_{\text{i}}^{\text{V}}
\label{eq_query}
\vspace{-1pt}
\end{equation}

\noindent Then, we apply two cross-attention blocks in parallel, each made of a one head scaled dot-product attention operation, with residual connection.  We obtain the SF context vector $h_{\text{SF}}$ by using the SF feature vector as a query and the SSL feature vector as key and value, and vice versa to obatain $h_{\text{SSL}}$, the SSL context vector.
Eq.~\eqref{eq_att_score1} and Eq.~\eqref{eq_att_score2} describe those symetric attention mechanisms, where $\cdot$ is the dot-product operator.
\vspace{-5pt}
\begin{equation}
    h_{\text{SF}} = \textsc{SoftMax}(\frac{Q_{\text{SF}}\cdot K_{\text{SSL}}}{\sqrt{D}})V_{\text{SSL}} + f_{\text{SF}}(S)
    \label{eq_att_score1}
\end{equation}
\vspace{-5pt}
\begin{equation}
    h_{\text{SSL}} = \textsc{SoftMax}(\frac{Q_{\text{SSL}}\cdot K_{\text{SF}}}{\sqrt{D}})V_{\text{SF}} + f_{\text{SSL}}(S)
    \label{eq_att_score2}
\end{equation}
\vspace{-5pt}

\noindent Our final feature is a projection on $\mathbb{R}^D$ of the concatenation of $h_{\text{SF}}$ and $h_{\text{SSL}}$, as descibed in Eq.~\eqref{eq_last_coatt}, where $\mathbin\Vert$ design the vector concatenation operation.
\vspace{-2pt}
\begin{equation}
    f_{\text{FUSE}}(S) = \textsc{Linear}( h_{\text{SF}} \mathbin\Vert h_{\text{SSL}})
    \label{eq_last_coatt}
    \vspace{-2pt}
\end{equation}

\noindent We also designed an attention-based fusion, however performance on preliminary experiments were weak compared to the co-attention model. We assume that the parallel computations on SF and SSL enable more sophisticated combinations of the two feature extractors than only one attention block would do. 

\vspace{-4pt}
\subsection{Mixture of Experts \label{moe_sec}}
\begin{figure}[!ht]
    \centering
    \includegraphics[scale=.30]{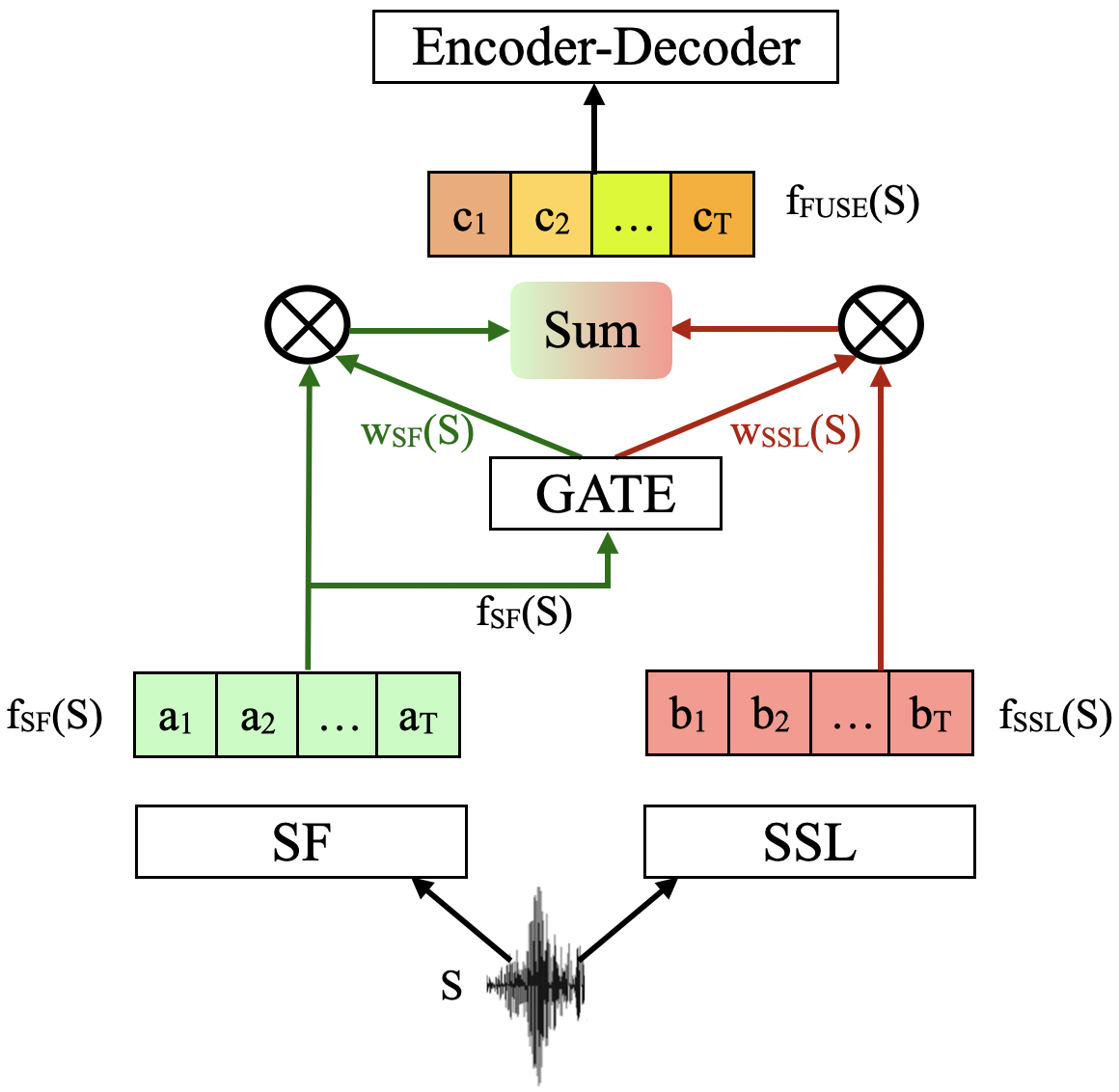}
    \caption{Architecture of the model combining SF and SSL through a gating mechanism. For a given utterance, the features are extracted by the two models ($a_{i}$ for SF and $b_{i}$ for SSL, $i\in \{1,...,T\}$).  Each model gets confidence scores and features are then summed. The $c_{i}$ variables indicates the weighted sum. Colors of $c_{i}$ frames are used to show how each frame gets a specific combination of SF (green) and SSL (red) features.}
    \label{fig:architecture_moe}
    \vspace{-15pt}
\end{figure}

\vspace{-2pt}
To get a broader understanding of the potential complementarity of SF and SSL features, we propose an adaptation of the mixture of experts \cite{MoE} gating paradigm, illustrated in Fig.~\ref{fig:architecture_moe}. We consider the two feature extractors, $f_{\text{SF}}$ and $f_{\text{SSL}}$, as our experts. This model requires a same number of frames for the two experts (see the processing step in Sec.~\ref{sec_feature_extraction}). We use $f_{\text{SF}}(S)$ as input feature to the gate.\footnote{Both $f_{\text{SF}}(S)$ or $f_{\text{SSL}}(S)$ could be used as input for the gate layer. We discuss this designing choice in Sec~\ref{ssec: experimental setup}.} 
Weights are calculated following Eq.~\eqref{eq:weights_moe}, where $w(S)\in  \mathbb{R}^{T\times 2}$ is the obtained weight matrix.

\begin{equation}
    w(S) = \Theta (f_{\text{SF}}(S)W_{\text{MoE}}), 
    \label{eq:weights_moe}
\end{equation}

\noindent with $W_{\text{MoE}} \in  \mathbb{R}^{D\times 2}$ a learnable matrix, and $\Theta (\cdot) $ a gating-type function such as \textsc{SoftMax}. \vspace{2pt}
For clarity, we introduce $w_{\text{SF}}(S), \: w_{\text{SSL}}(S) \in \mathbb{R}^{T} $, the column vectors of $ w(S)$.

\noindent The final combined feature is computed following Eq.~\eqref{eq:moe_final}, where $[x]^{\text{tr}}$ denotes the transpose vector of $x$. 
\begin{equation}
 \vspace{-2pt} 
    f_{\text{FUSE}}(S) = \sum_{i\in \{\text{SF} , \text{SSL}\}} [w_\text{i}(S)]^{\text{tr}} f_{\text{i}}(S) 
    \label{eq:moe_final}
    \vspace{-2pt} 
\end{equation}

\noindent The mixture of experts model outputs a weighted sum of feature extractors for each frame of the utterance. The weights can be interpreted as confidence scores of SF and SSL for each frame. This model makes the fusion process more interpretable by enabling to compare relative usage of SF and SSL.

\section{Experiments}
\vspace{-3pt}
\subsection{Datasets}
\vspace{-2pt}
Totonac is an endangered language spoken in the northern sierras of the state of Puebla and adjacent areas of Veracruz, Mexico. To increase the coverage over endangered languages, we evaluate our proposed methods on Totonac and release a publicly available version of Totonac ASR data.\footnote{\url{http://www.openslr.org/107/}} The corpus comprises 10 hours of speech (86 long recordings) with fine-grained transcriptions. We randomly selected 70 recordings for the training set, 8 for validation, and 8 for testing.\footnote{Those splits are officially released at \url{https://github.com/ftshijt/Totonac_Split.git}}
In addition to Totonac, we perform experiments on Arabic corpora of 20 hours from Commonvoice 5.1 \cite{Ardila2020CommonVA}, still in the low-resource scenario. Finally, we extend our study to low resource ST using the Mboshi-French dataset \cite{DBLP:mboshi}, consisting  of 4 hours of speech, to show that our framework is effective in other speech tasks as well. We chose Arabic (Semitic language) and Mboshi (Bantu language) as they belong to different language groups than English (Germanic). Thus, we will compare the robustness of the SSL representations to the ones of our proposed models over a set of diverse language families, all different from the one of the SSL self-training data.

\subsection{Experimental setup}

\label{ssec: experimental setup}

\par \textbf{Baseline :} Our ASR baseline (\textbf{Base} in the experiments) adopts a transformer-based encoder-decoder architecture with CTC/Attention hybrid training \cite{kim2017joint}. The front-end extracts FBANK spectral features with a frame length of 25ms and a frame-shift of 10ms. The extracted FBANK features are  subsampled with a convolutional block and then fed into the encoder-decoder. The encoder consists of 12 self-attention blocks with 4-head attention and 256-dimensional hidden sizes while the decoder has 6 cross-attention transformer blocks. For ST, we add 2 extra decoders of 2 layers each to this architecture. SpecAugment \cite{park19e_interspeech} and speed perturbation are employed for data augmentation. Hyperparameters used for training can be found on \href{https://github.com/espnet/espnet}{ESPnet}. The ASR model is trained to recognize 250 byte-pair-encoding (BPE) units. The same architecture and training configuration are used for the following experiments.
\vspace{0.25em}

\noindent \textbf{Self-supervised representations :} In our experiments, we employ HuBERT \cite{DBLP:HuBERT} , which shows promising results over the SUPERB benchmark \cite{tsai2022superbsg}.\footnote{We also performed preliminary experiments over Wav2vec2 XLSR model \cite{DBLP:XLSR}, but it did not improve the results over HuBERT model so we continued the study only for HuBERT model.} To fully explore the potential of HuBERT, we select the HuBERT-large model pre-trained over 60k hours of LibriLight \cite{librilight, ott2019fairseq}.
The SSL wrapper provided in Yang et al.\cite{yang21c_interspeech} is applied to extract high-dimensional features with a 20ms frame-shift.  
In experiment \textbf{SSL}, the FBANK feature extractor (used in \textbf{Base}) is replaced by the pretained HuBERT model, which is fine-tuned during training.\footnote{SSL based front-ends could be freezed, but the best performances were obtained when fine-tuning the models.} 

\vspace{0.25em}
\noindent \textbf{Learnable combinations :} Experiments \textbf{Linear, Conv.} and \textbf{co-Att.} are the \textsc{Transform} operations introduced in Eq.~\eqref{eq_transform} of Sec~\ref{sec_feature_comb} respectively for concatenation, convolutional, and co-attention based fusions. For \textbf{Linear} experiment, we concatenate $f_{\text{SF}}(S)$ and $f_{\text{SSL}}(S)$ and then project the concatenation into a 80-dimensional space. In \textbf{Conv.} experiment, we apply a 1-dimensionnal convolutional layer with kernel size 5 and stride of 1 over $f_{\text{SF}}(S)$ and $f_{\text{SSL}}(S)$ before concatenating and projecting them. The co-attention model is described through Eq.~\eqref{eq_query} to Eq.~\eqref{eq_last_coatt}, and the model is illustrated in Fig.~\ref{fig:architecture_coatt}. 

\vspace{0.25em}

\noindent \textbf{Mixture of experts :} Our mixture of experts model (\textbf{MoE} in the experiments) follows Eq.~\eqref{eq:weights_moe} and Eq.~\eqref{eq:moe_final} described in Sec.~\ref{moe_sec}. For the main experiments, we use SF (here FBANK) as input features and $\Theta(\cdot)=\textsc{Log-SoftMax}(\cdot) $ for the gating function. 
We performed a comparative study of inputs to the gating function. Using both SF or SSL features led to better scores than the baselines but SF as input performed best. Our interpretation is that it is easier for the model to learn gating weights when computed over non-learnable features (SF, here FBANK) than over complex features which are continuously fine-tuned.
We also compared results with $\Theta(\cdot)=\textsc{Log-SoftMax}(\cdot) $ and $\Theta(\cdot)=\textsc{SoftMax}(\cdot) $. Performances are similar, $\Theta(\cdot)=\textsc{Log-SoftMax}(\cdot) $ being slightly better. A more detailed analysis of the gating weights intra-utterance revealed a more peaky behavior for $\Theta(\cdot)=\textsc{SoftMax}(\cdot) $, which in our opinion led to the small performance degradation.

\vspace{0.25em}

\noindent \textbf{Evaluation metrics :} We use Character Error Rate (CER) for evaluation of our ASR models and BLEU score to measure performances of our ST systems.

\section{Results and Analysis}

\subsection{Main results}

Table~\ref{tab_res} provides results for the experiments listed in Sec.~\ref{ssec: experimental setup}.

\begin{table}[ht!]
  \centering

    \small
    \caption{ASR and ST results over models described in Sec.~\ref{ssec: experimental setup}. The two first experiments are our FBANK and SSL baselines. The following lines are the proposed Linear, Convolutional, co-Attention, and Mixture of Experts models.}
    \vspace{-8pt}
\small
\begin{tabular}{c|c|c||c}

\toprule
&  \multicolumn{2}{c}{CER $\downarrow$}  & BLEU $\uparrow$ \\
\midrule
Exp  & Totonac & Arabic & Mboshi-French \\
\midrule
\textbf{Base} & 17.2 & 15.4  & 10.9 \\
\textbf{SSL}& 14.2 & 8.1  & 10.6\\
\midrule
\textbf{Linear} & 14.0 & 6.6   & \textbf{11.6} \\
\textbf{Conv.} &  13.9 & 7.2 & 11.3 \\
\textbf{co-Att.} &  \textbf{13.4} & \textbf{5.4} & 10.9\\
\textbf{MoE} & 13.7 & 6.2 &  11.2  \\

\bottomrule

\end{tabular}

  \label{tab_res}
  \vspace{-12pt}
\end{table}

\subsubsection{Speech Recognition results}
First we remark that using HuBERT as a feature extractor (\textbf{SSL} experiment) instead of FBANK (\textbf{Base} experiment) is very effective on the Totonac and Arabic ASR corpora, leading to respective diminutions of $3.0$ and $7.3$ of CER. 

\noindent Then, we note that all of the combination methods we introduced in Sec.~\ref{sec_feature_comb} led to improvements on the two datasets over the \textbf{Base} and \textbf{SSL} models. We get a diminution of $2.7$ CER ($33\%$) on Arabic and $0.8$ CER ($5.6\%$) on Totonac when using the co-attention model. The co-attention model performs better than the linear and convolution based methods, in particular for Arabic. A possible explanation is that this model: (1) has a larger modeling capacity (leading to better results), and (2) induces a more balanced use of the two front-ends, through the symmetric architecture and the residual connections. This second point could explain the greater CER reduction over Arabic than Totonac, as an equal contribution of front-ends seems to lead to a robust model for Arabic (see Sec.~\ref{sec_abl_moe}).

\noindent Finally, the mixture of experts model that we introduced for gaining interpretability is also getting strong performances. 

\subsubsection{Speech Translation results}

As it is straightforward to use our front-end fusion framework for different speech tasks, we applied it to ST. Table~\ref{tab_res} shows that all of our proposed methods outperforms both FBANK and SSL baselines. We note that using HuBERT representations as front-end degraded the performance in that scenario (see experiments \textbf{Base} and \textbf{SSL}). Even in that case, all the proposed systems performed better than both baselines. The linear fusion method reaches a BLEU score of $11.6$, gaining $1.0$ BLEU over the SSL baseline and $0.7$ BLEU over the FBANK one. Contrary to the ASR scenario, here the linear and convolutional methods outperform the co-attention one. As Mboshi is only made of only 4 hours of speech, we assume that the co-attention model may be too complex to be well trained contrary to the linear model.

\subsection{Mixture of Experts : Weights and Analysis \label{sec_abl_moe}}

In this section, we examine the weights $w_{\text{SF}}(S)$ and $w_{\text{SSL}}(S)$ (introduced in Sec.~\ref{moe_sec}) obtained by the mixture of experts model for the two ASR datasets.
For more interpretability, we normalized them so that $w_{\text{SSL}}(S)+w_{\text{SF}}(S)=1$. First we can note that our robust \textbf{MoE }model is indeed using both FBANK and HuBERT components as the two weights are non negligible. Then, we remark that the weights across frames of a same utterance are quite similar. The two front-ends are used consistently over the frames, which we would expect as a utterance content may be quite consistent. We note that the weights across different utterances are also similar within languages. However they are very different from one language to another.

\begin{table}[ht!]
    
  \centering
    \caption{Two views on HuBERT representations quality over Totonac and Arabic data. The first column presents $\overline{w_{\text{SSL}}(S)}$, the mean \textbf{MoE} weights for HuBERT front-end. The second column is the character error reduction rate reduction (CERR\footref{fn:refnote}) between the FBANK baseline and the HuBERT baseline.}
    \vspace{-3pt}
    \begin{tabular}{c|cc}
\toprule
Language & $\overline{w_{\text{SSL}}(S)}$ & CERR(\textbf{Base} $\rightarrow$ \textbf{SSL})  \\
\midrule
Totonac & 0.17 & 17\%   \\
Arabic & 0.51 & 47\%  \\
\bottomrule
\end{tabular}
    \vspace{-9pt}
  \label{weigth_table}
\end{table}

\noindent The first column of Table~\ref{weigth_table} presents the mean $w_{\text{SSL}}(S)$ weight for each language. Contrary to the Arabic model, which uses HuBERT and FBANK with similar weights, the Totonac model seems to be using HuBERT representations as an adjustment component, relying on average at more than $80\%$ on spectral features.
Our interpretation is that the Commonvoice Arabic data is closer in domain (read speech) to the English LibriLight than Totonac data is (spontaneous speech/conversation). For that reason, HuBERT model may extract relatively better speech representations (compared to FBANK representations) for Arabic than it does for Totonac. This would explain that the mixture of experts model grants HuBERT with a larger weight for the Arabic data. Another way of quantifying HuBERT representations quality over the languages could be to calculate the character error reduction rate (CERR\footnote{CERR is defined as follows : CERR = $\frac{\text{CER(Base) - CER(SSL)}}{\text{CER(Base)}}\times 100$.\label{fn:refnote}}) between FBANK and HuBERT baselines (experiments \textbf{Base} and \textbf{SSL} in Table~\ref{tab_res}). The second column in Table~\ref{weigth_table} confirms our intuition : the mixture of experts model weights the components according to their relative strength over the language. As Arabic benefits more from HuBERT representations than Totonac does, the mixture of experts model assigned a higher weight to the HuBERT front-end in the Arabic model than in the Totonac one.

\section{Conclusions}

SSL models performance depends highly on the relatedness between the self-supervised training  domain(s) and the target data domain. As spectral features are not subject to those variations, we proposed a framework to combine spectral features to SSL representations. This framework can be applied to many speech tasks with no further work. We obtained strong improvements over ASR and ST datasets compared with the SSL baseline. We further proposed a weight analysis showing that: (1) our models performances are strong for both in-domain and out-of-domain scenarios. (2) our mixture of experts framework enables quantifying the domain shift between the SSL training data and the target language resources.

\noindent Future work could involve fusions at the encoder level. As SSL models also perform strongly when used as encoders, fusing SSL features with SF passed through a pre-trained encoder could be an even more robust technique.

\section{Acknowledgements}

This work used the Extreme Science and Engineering Discovery Environment (XSEDE) ~\cite{xsede}, which is supported by National Science Foundation grant number ACI-1548562. Specifically, it used the Bridges system ~\cite{nystrom2015bridges}, as part of project cis210027p, which is supported by NSF award number ACI-1445606, at the Pittsburgh Supercomputing Center (PSC). Recording (Amith) and transcription (López) of Zongozotla Totonac was carried out with support from the National Science Foundation, Documenting Endangered Languages Program (Award \#BCS-1401178), the National Endowment for the Humanities, Preservation and Access (Award \#PD-50031-14), and the Jacobs Research Fund (awards in 2019 and 2020). Amith was PI on all three.

\bibliographystyle{IEEEtran}

\bibliography{mybib}

\end{document}